\documentclass[conference]{IEEEtran}
\IEEEoverridecommandlockouts
\usepackage{cite}
\usepackage{url}
\usepackage{amsmath,amssymb,amsfonts}
\usepackage[linesnumbered,ruled]{algorithm2e}
\usepackage{graphicx}
\usepackage{textcomp}
\usepackage{xcolor}
\def\BibTeX{{\rm B\kern-.05em{\sc i\kern-.025em b}\kern-.08em
    T\kern-.1667em\lower.7ex\hbox{E}\kern-.125emX}}

\begin{document}

\title{Identifying phase transitions in physical systems with neural networks: a neural architecture search perspective \\
}

\author{\IEEEauthorblockN{1\textsuperscript{st} Rodrigo Carmo Terin}
\IEEEauthorblockA{\textit{Intelligent Systems Group} \\
\textit{University of the Basque Country}\\
San Sebastian, Spain \\
rodrigo.carmo@ehu.eus}
\and
\IEEEauthorblockN{2\textsuperscript{nd} Zochil Gonz\'alez Arenas}
\IEEEauthorblockA{\textit{Department of Applied Mathematics} \\
\textit{University of the State of Rio de Janeiro}\\
Rio de Janeiro, Brazil \\
zochil@ime.uerj.br}
\and
\IEEEauthorblockN{3\textsuperscript{rd} Roberto Santana}
\IEEEauthorblockA{\textit{Intelligent Systems Group} \\
\textit{University of the Basque Country}\\
San Sebastian, Spain \\
roberto.santana@ehu.eus}
}

\maketitle

\begin{abstract}
 The use of machine learning algorithms to investigate phase transitions in physical systems is a valuable way to better understand the characteristics of these systems. Neural networks have been used to extract information of phases and phase transitions directly from many-body configurations. However, one limitation of neural networks is that they require the definition of the model architecture and parameters previous to their application, and such determination is itself a difficult problem. In this paper, we investigate for the first time the relationship between the accuracy of neural networks for information of phases and the network configuration (that comprises the architecture and hyperparameters). We formulate the phase analysis as a regression task,  address the question of generating data that reflects the different states of the physical system, and evaluate the performance of neural architecture search for this task. After obtaining the optimized architectures, we further implement smart data processing and analytics by means of neuron coverage metrics, assessing the capability of these metrics  to estimate phase transitions. Our results identify the neuron coverage metric as promising for detecting phase transitions in physical systems. 
\end{abstract}

\begin{IEEEkeywords}
  phase transition, physical systems, smart data analysis, neural networks, neural architecture search
\end{IEEEkeywords}

\section{Introduction} \label{sec:INTRO}

 In statistical and condensed-matter physics it is essential the identification of phases of matter and their transitions which are associated to drastic changes in the macroscopic properties of the physical system. This process usually involves a data generation phase in which data from the system dynamics is obtained, followed by a posterior phase in which machine learning methods are applied to extract insights from this data.  An increasing number of works have investigated the use of neural networks to identify phases and phase transitions in physical systems~\cite{Ng_and_Yang:2023,Tian_et_al:2023,Van-Nieuwenburg_et_al:2017} from data. Using state configurations sampled from the system at different phases as inputs, these models have shown accurate results to detect multiple types of order parameters~\cite{Carrasquilla_and_Melko:2017}.
 The theory of phase transitions was developed in the context of equilibrium statistical mechanics. More recently, out-of-equilibrium systems and quantum physics have brought new approaches to phases, phase transitions, and criticality. In this context, dynamic and topological phase transitions have been studied, where local order parameters do not exist. For these systems, machine learning techniques also appear as a promising approach for studying phases and finding non-conventional order parameters~\cite{Ni_et_al:2019,Che_et_al:2020,Kuliashov_et_al:2023}.

 Phase detection can be approached using supervised and unsupervised machine learning approaches. Neural networks have been used as part of both approaches. In the supervised approaches~\cite{Tian_et_al:2023}, the input-label pairs are usually formed by the system configuration and the parameter (or parameters) from which the configuration has been sampled. The neural network then learns to predict the parameter, and some measure of the change in quality of the approximation for the predicted parameters is employed to detect a possible phase transition.  For the unsupervised approximations~\cite{Ng_and_Yang:2023,Tanaka_and_Tomiya:2017,Wang:2016}, no prior labeling of the dataset is required and the patterns are exclusively extracted from the inputs. One example on the application of neural networks in the unsupervised approach is learning with  autoencoders, and the subsequent analysis of the reconstruction error as the order parameter changes~\cite{Ng_and_Yang:2023}. 

 Two questions that have been traditionally overlooked in the research on the application of neural networks to analyze phases and phase transitions are: 1) To what extent the architecture of the network influences the prediction accuracy for the target property being investigated? 2) Can the internal patterns of activation of the neural networks for different input data  reveal information about the phase transition? In this paper we investigate these two questions.

 To analyze the influence of the network architecture, we employ a neural architecture search (NAS)~\cite{Elsken_et_al:2019,White_et_al:2023} strategy, in which the neural networks are optimized by a neuroevolutionary algorithm to minimize some error related to the order parameter prediction. By explicitly searching for high-performing neural networks, we can determine to what extent is the architecture design related to the accuracy in the prediction of the physical system parameters. 
 
 To determine whether there are different patterns of activation related to the phases of the physical system, we compute neuron coverage metrics~\cite{Ma_et_al:2018,Yang_et_al:2022} which have been conceived to quantify how the neural network components are covered by test instances. We would like to discern whether system configurations sampled from different order parameters determine different patterns of activation in the neurons (and/or) layers of the networks, and identify if any of the neuron coverage metrics is more correlated to these parameters. To the knowledge of the authors, coverage metrics have been not used before as a way to study the behavior of neural networks that predict phase transitions. 
 
 The paper is organized as follows: In the next section, some necessary background on phase transitions in the Ising model and  neuron coverage metrics are presented. This section also discusses works on the use of machine learning to investigate phase transition that is related to our proposal. In Section~\ref{sec:MODEL_NN}, the supervised machine learning scenario considered in this paper is described and the procedure to generate the synthetic data is explained together with the NAS strategy to search for high-performing neural networks for order parameter prediction. Section~\ref{sec:EXPE} introduces the experimental framework and presents the numerical results with a discussion of our findings. The conclusions of the paper are presented in Section~\ref{sec:CONCLU}, together with a discussion of numbered research directions for future work.


\section{Background}  \label{sec:BACKGROUND}

\subsection{Phase transitions in the Ising model}

A Hamiltonian function serves to describe how the energy of a physical process depends on the state of the system's particles and their interactions. The generalized Ising model is a theoretical model in statistical physics  
which describes the ferromagnetism phenomenon through a system of magnetic particles. It is a paradigmatic model for studying phase transitions given that its exact solution is known for one and two dimensions. 
  In the absence of an external magnetic field, it is described by the Hamiltonian shown in Equation~\eqref{ISINGM} where $L$ is the set of sites called a lattice. To each magnetic particle in the lattice, it is assigned a magnetic moment that is represented by a spin variable $\sigma_i$ at site $i \in L$ which either takes the value $1$ or $-1$. The former value represents the parallel and the latter the anti-parallel states. 
  A specific choice of values for the  spin variables is called a configuration. The constants $J_{ij}$  are the  interaction or couplings coefficients. The ground state is the  configuration that has all magnetic moments aligned, responsible for minimizing the energy.  
  
 \begin{equation}  
 H = - \sum_{i<j \in L} J_{ij} \sigma_i \sigma_j\,. 
 \label{ISINGM}  
 \end{equation}  
 The order parameter is a quantity that characterizes the phase of the system, it is zero-valued in one phase, and non-zero value in the other one. In the Ising model, the order parameter is the magnetization, which is computed as the average value of the spin configuration.
  The phase transition in the context of the Ising model is a transition from an ordered state where most of the spins are aligned in the same direction 
  below the critical temperature $T_{c}$, to a disordered state where there is a random distribution of the spin values above $T_{c}$, or vice versa. In addition, $T_c = 2.26 \, \text{K}$ in a theoretical viewpoint, where $\text{K}$ denotes the Kelvin unit. In this paper, we consider a temperature range of $1.0 \, \mathrm{K}$ to $3.5 \, \mathrm{K}$ for the Ising problem and address the prediction of the phase transition within this interval. Also, the value of the $T_{c}$  is related to the coupling coefficients. Spin couplings can take values from a priori defined set of values and both the choice of the set and the distribution with which the couplings are sampled from it influence the characteristics of the problem and its difficulty for the optimizers. 
 
It is important to note that acquiring real data for the Ising model is challenging due to experimental constraints and the difficulty in controlling all relevant physical parameters. As a result, simulated data provides a clearer insight into phase transitions, offering clean and controlled conditions in comparison with the experimental setup. Furthermore, understanding these transitions within the Ising model sheds light on critical phenomena and collective behaviors in ferromagnetic materials \cite{uzunov1993introduction}. Accurate phase detection is essential for predicting a material's properties across the range of temperatures. This is directly related to temperature prediction, as the material's phase is significantly influenced by temperatures relative to the critical threshold $T_c$.

\subsection{Machine learning models for the analysis of phase transition}

Machine learning algorithms have been increasingly applied to study phase transition.  Approaches to order parameter prediction and phase transition detection can be grouped into three different type of approaches: 1) Supervised machine learning; 2) Unsupervised machine learning; and 3) Learning-by-confusion. We briefly discuss these approaches since they are very related to our contribution.


In the supervised machine learning approach, data obtained from physical systems (e.g., spin configurations) can be assigned a class (e.g., a discrete value indicating a set of values of the order parameter, or a given phase), or directly assigned a target variable (e.g., the temperature of the system). Then, a machine learning model is trained to learn this mapping between inputs and classes for classification  problems, or the mapping between inputs and the target variable for regression problems.


In unsupervised machine learning approaches, no labels about the order parameters or any other parameter that describes the state of the system are used. In this strategy, a characterization of the system can be extracted from some latent variables learned or from some patterns extracted from data. For example, in~\cite{Wetzel:2017}, Wetzel analyzes  a number of unsupervised machine learning methods and concludes  that principal component analysis and variational autoencoders ~\cite{Kingma_and_Welling:2013} are the most promising ones to reveal phase transitions. In~\cite{Wang:2016}, unsupervised learning algorithms are applied to identify phase transitions in Ising models. 


The learning-by-confusion approach \cite{Van-Nieuwenburg_et_al:2017} involves training classifiers to differentiate between two phases of a system at various points along a parameter grid. It focuses on how the classifier's performance, which is in general measured by uncertainty or entropy, varies across this grid, with special attention to the classifier's behavior near the critical points of phase transitions, where confusion (or uncertainty) is commonly maximal. Arnold et al.~\cite{Arnold_et_al:2023} have recently proposed to reduce the cost  of the confusion method by using a single neural network that solves a multi-class classification task (in~\cite{Arnold_et_al:2023}, this approach is called \emph{Multi-Task Learning-by-Confusion}).

\subsection{Neuron coverage}  \label{sec:COV_METRICS}

 In the literature, there are slightly different definitions of the neuron coverage metrics. We have mainly adopted the conventions used in~\cite{Ma_et_al:2018,Yang_et_al:2022}, with some few changes. 
 
 Let $c$ represent a neuron of a multi-layer perceptron (MLP) of $A$ layers.  $M_1, M_2, \dots, M_A$ represent the number of neurons in each layer and  $N = \sum_{j=1}^{A} M_j$ is the total number of neurons in the network.
 
 We use $\phi(x^i,c)$ to denote the  function that returns the output of neuron $c$ given $x^i$ as input. For a given neuron $c$,  it is said to be \emph{activated} for a given input $x^i$ if  $\phi(x^i,c)>t$,  where $t$ is a given threshold. $L_{c}$ and $H_{c}$ will respectively represent the lower and upper bounds of function  $\phi(x^i,c)$ for $x^i \in D$. These values are determined by analyzing the values of $\phi(x^i,c)$ for the training dataset $D$. Usually, the set $D$ corresponds to a set of instances in the training dataset, i.e., $D=D_{train}$.

  \subsubsection{Neuron coverage metrics}

  Given a set of instances $D$ and a given threshold $t$, the neural network coverage~\cite{Pei_et_al:2017} measures the proportion of neurons in MLP that have been activated by at least one instance in $D$:

   \begin{equation}
    NC = \frac{ \left| \{c| \,  \exists x^i \in D: \phi(x^i,c) > t \} \right|} {N}
   \end{equation}

 \subsubsection{Top-$K$ neuron coverage}
For a given test input $x^i$ and neurons $c$   and $c'$ in the same layer, $c$ is more active than $c'$ if  $\phi(x^i,c) > \phi(x^i,c')$.  For the $j$-th layer, $top^j_K(x^i)$ on layer $j$  denotes the set of neurons that have the largest $K$ outputs on that layer given $x^i$.

The top-$K$ neuron coverage (TKNC) measures how many neurons have once been among the most active $K$ neurons on each layer.

  \begin{equation}
    TKNC(D_{test},K) =  \frac{|\bigcup_{x^i \in D_{test}} (\bigcup_{1 \leq j \leq A} top^j_K(x^i)))|} {N}
  \end{equation}

 \subsubsection{k-multi-section neuron coverage}
 
 Given a neuron $c$,  the multi-section neuron coverage measures how thoroughly the given set of test instances  covers the range $[L_{c},H_{c}]$. The range is divided into $k>0$ equal sections,  called multi-sections.  A multi-section $S^s_{c}, \; s \in \{1,\dots,k\}$ is said to be covered if $\phi(x^i,c) \in S^s_{c}$ for $x^i \in D_{test}$.

 The  k-multi-section neuron coverage for neuron $c$ is defined~\cite{Ma_et_al:2018} as the ratio between the number of sections covered by $D_{test}$ and $k$, 

 \begin{equation}
    \overline{KMN}(c) = \frac{ \left | \{S^s_{c}| \, \exists x^i \in D_{test}: \phi(x^i,c) \in S^s_{c} \} \right|}{k}
 \end{equation}
 
 The k-multi-section neuron coverage of an MLP \cite{Ma_et_al:2018} is defined as:

  \begin{equation}
    KMN(D_{test},k) =  \frac{ \sum_{c}   \overline{KMN}(c) }{k \cdot N }
 \end{equation}

\subsubsection{Neuron boundary coverage and strong neuron activation coverage}
     
  A test input $x^i \in D_{test}$ is said to be located in the corner-case region of an MLP if there is a neuron $c$ such that $\phi(x^i,c)$ is lower than $L_{c}$  or higher than $H_{c}$. 

  To cover corner-case regions of MLPs, the sets of covered corner-case regions are defined as: 

  \begin{align}
     LCN &=&  \{c| \, \exists x^i \in D_{test}: \phi(x^i,c) \in (-\infty,L_{c}) \}\\
     UCN &=&  \{c| \, \exists x^i \in D_{test}: \phi(x^i,c) \in (H_{c},+\infty) \}
  \end{align}

  The neuron boundary coverage (NBC) measures how many corner-case regions have been covered by the given test input set $D_{test}$. 

  \begin{equation}
    NBC(D_{test}) =  \frac{|LCN|+|UCN|}{2 \cdot N }
  \end{equation}

  The strong neuron activation coverage (SNAC) measures how many corner cases, with respect to the upper boundary value, have been covered by the given test inputs $D_{test}$.
  
  \begin{equation}
    SNAC(D_{test}) = \frac{|UCN|}{N}
  \end{equation}

 \subsubsection{Implementation}

We have used the implementation of the neuron coverage metrics\footnote{Available from \url{https://github.com/DeepImportance/deep\\importance_code_release}} developed as part of the work presented in~\cite{Gerasimou_et_al:2020}.

 \subsection{Related work}

\subsubsection{Neural networks for phase transition}

Tian et al~\cite{Tian_et_al:2023} proposed a supervised approach in which some descriptor of the spin configurations are first extracted and then used as inputs to the neural network in order to predict the order parameter. As loss function, the authors propose the use of mean squared error (MSE) or cross-entropy and select the two-dimensional Ising model as a benchmark system. 

Ng and Yang~\cite{Ng_and_Yang:2023} proposed a modified method of anomaly detection for phase characterization within the framework of unsupervised machine learning. They use the reconstruction error measured as the MSE of a convolutional autoencoder and study phase transitions in various spin models defined on a square lattice. The authors highlight that the standard deviation of the MSE can be a better indicator for phase transitions compared to MSE alone. 

Beyond purely supervised and unsupervised approaches, semi-supervised methods combine labeled and unlabeled data to learn more accurate models. In~\cite{Chen_et_al:2023}, a domain-adaptation algorithm is proposed to investigate the phase transition of the two-dimensional q-state Potts models on the square lattice with nearest-neighbor interactions. The domain-adversarial neural network~\cite{Ganin_et_al:2016} is used for this purpose. 

One of the limitations of the initial learning-by-confusion approaches was the need to learn multiple models to solve the same type of machine learning task but for different ranges of parameters.  As an alternative, Liu and VanNieuwenburg \cite{Liu_and_VanNieuwenburg:2018} proposed an unsupervised machine-learning scheme for detecting phase transitions that uses only two different  discriminative cooperative networks (DCNs) which cooperate to detect phase transitions from fully unsupervised data. Recent works have proposed other strategies to avoid the use of the multiple models \cite{Arnold_et_al:2023}. The Learning-by-confusion approach has also been recently applied to the identification of discontinuous phase transitions~\cite{Richter_et_al:2023}.

\subsubsection{NAS and phase transitions}

  Research on NAS is an active field in machine learning and a variety of methods have been proposed in the area~\cite{Elsken_et_al:2019,White_et_al:2023}. Particularly related to our work are neuro-evolutionary approaches~\cite{Floreano_et_al:2008,Stanley_et_al:2019} that use evolutionary algorithms to search the space of architectures. In this paper, we use a neural architecture representation similar to the one introduced in~\cite{Garciarena_et_al:2018}  in which the neural network configuration is encoded as a list. The evaluation of each network configuration implies the creation of a network architecture by decoding the specification in the list, and learning the parameters of the network from data using a variant of batch gradient descent. 

  We have not found any reference to the use of NAS for finding high-performing neural networks for the order parameter prediction and phase detection results. However, the question of the influence of the neural network structure for the problem of identifying the phase transition in the Ising model has been previously addressed. In~\cite{Kim_and_Kim:2018},  the authors conclude that having two hidden neurons in the neural network architecture can be enough for an accurate prediction of critical temperature in the Ising model. In a similarly surprising result, when comparing shallow and deep learning architectures to approximate the probability distribution of a two-dimensional  Ising system near criticality,  Morningstar and Melko \cite{Morningstar_and_Melko:2018} conclude that shallow networks are more efficient than deep networks at representing this probability. 

\section{Phase parameter prediction using neural networks} \label{sec:MODEL_NN}

  In this section, we present the different steps involved in the use of the neural networks for estimating the tuning parameters of the physical system, and from these parameters to study the phase transition.  We focus on the Ising model defined on a grid of $8 \times 8$ dimension with $n=64$, and we consider this problem for a set of $26$ temperature values ranging from $1.0 \, \mathrm{K}$ to $3.5 \, \mathrm{K}$ as mentioned in Sec. \eqref{sec:BACKGROUND}. The regression problem to be solved by the neural network is to predict the temperature at which a given configuration of the systems has been sampled. 
  The process is organized according to the following steps:
  
 \begin{enumerate}
     \item Generating samples from the Ising configuration at different temperatures.
     \item Training a neural network to predict the temperature using a (training) subset of the generated data.
     \item Evaluating the performance of the neural network using a (test) subset of the generated data.
     \item Evaluating the internal behavior of the neural network for data corresponding to different temperatures by means of the neuron coverage metrics. 
 \end{enumerate}

\subsection{Procedure for data generation}


In this work, we have used the Markov chain Monte Carlo method to build up our dataset. The central part of our construction consists on using two distinct spin update algorithms, namely: the Metropolis and Wolf cluster updates for simulating the spin magnetic system \cite{PhysRevLett.62.361}.

Initially, we defined the model parameters: the size of the spin grid, which is equal to eight; the range of the temperature; and the number of steps in the simulation for each temperature, set at $5000$. The equilibration steps played a pivotal role, defined as $512$ iterations of the Metropolis algorithm before collecting data to guarantee a representative state of thermal equilibrium for each temperature presented in the range and minimizing the effects of random initial configurations. The decorrelation steps, equal to the grid size, are used to reduce sample correlation. Finally, we set $J=1.0$ to represent the ferromagnetic energy scale.

In order to generate the $2D$ classical Ising model dataset, we worked with four different functions. The first one is defined as Metropolis Update Function responsible for updating a single state of the spin in the grid. A spin configuration is randomly chosen and its energy variation is computed based on spin`s interaction with neighborhood and the interaction couplings $J$. The spin changes its sign when either the variation lowers the system´s energy or by a probability determined by the temperature and the energy variation.

After that, we used the Wolff Cluster Update Function in charge of updating a cluster of spins. It forms a cluster that contains the spin randomly selected before together with similarly aligned spins. The probability of adding such neighboring spins depends on the temperature and the coupling parameter $J$. All the spins inside the cluster change their sign collectively turning out easy to rearrange them and speeding up our dataset generation.

The third one is the Markov Chain Simulation responsible for simulating the aforementioned system for a given temperature. The procedure starts by running the Metropolis update in an equilibrium state with the already defined number of steps in order to bring the system to a near equilibrium phase and then for each simulation steps it alternates among saving the current spin configuration and updating the system through the Wolf cluster algorithm. After that, the spin configurations and the corresponding temperatures are stored for our analysis.

The last one is the Data Generation and Storage function in charge of simulating across all the range of temperatures, i.e., for each temperature, it starts with the random spin grid, then conducts the Markov chain procedure and finally collects the data. The latter contains spin configurations and their respective temperatures which are stored in a CSV file.


\subsection{A neuroevolutionary approach for finding high-performing architectures}

  Once the sets of the Ising model configurations have been generated using the procedure explained in the previous section, a multi-layer perceptron (MLP) neural network can be learned using as loss function the MSE. However, the design of the best network architecture for this problem is an open question. Therefore, we address this question as a neural architecture search problem in which we want to find the architecture that minimizes the prediction error for the temperatures. 

  To solve the optimization problem, we use a neuroevolutionary algorithm in which each possible neural network architecture, as well as other hyperparameters are represented as a list. Among the other hyperparameters that are optimized as part of the process are the type of weight initialization and activation functions. The weights of the networks are not codified. They are learned using a gradient optimization technique during the learning process. 
  

  During the fitness evaluation step, each architecture is trained using a subset of the training data, and its fitness is computed on a different validation set as the MSE. This is the fitness associated to the network. Once the evolutionary process has been completed, the test set is used to evaluate the performance of the networks that are present in the final population.  Algorithm~\ref{Alg:GA_DNN} shows the pseudocode of the neuroevolutionary algorithm.

\begin{algorithm}[ht]
	Set $t\Leftarrow 0$. Create a population $D_0$ by generating $N$ random MLP descriptions\;
	\While{halting condition is not met}{
		Evaluate $D_t$ using the fitness function\;
		From $D_t$, select a population $D_t^S$ of $K \leq N$ solutions according to a selection method\;
		Apply mutation with probability $p_m=1-p_x$ to $D_t^S$ and create the offspring set $O_t$. Choice of the mutation operator is made uniformly at random\;
		Create $D_{t+1}$ by using the selection method over $\{D_{t},O_{t}\}$\;
		$t \Leftarrow t+1$\;
	}
	\caption{Neuroevolutionary algorithm for parameter prediction.}
	\label{Alg:GA_DNN}
\end{algorithm}

To implement the neuroevolutionary approach we have used the \texttt{deatf} library \footnote{Available from \url{https://github.com/IvanHCenalmor/deatf}} implemented in tensorflow \cite{Abadi_et_al:2016} and that uses genetic operators originally included as part of the \texttt{DEAP} library \cite{Fortin_et_al:2012}. Further details about the neuroevolutionary algorithm can be found in \cite{Garciarena_et_al:2020,Santana_et_al:2023}. 



\subsection{Evaluating the behavior of high-performing architectures}

   
   The error on the approximation for the different values of the temperature can provide information of how the difficulty of the machine learning task changes according to the disorder in the spin configurations. Another important research objective of this paper is to understand how the internal behavior of the neural networks is influenced by Ising configurations generated at  different temperatures. This question is highly relevant in ML approaches to investigating phase transitions, as state-of-the art approaches typically focus on the predictions made by neural networks, rather than on any metric that captures the dynamics of the learning or inference steps. Our contribution to investigating this issue is based on the use of coverage metrics.

   For each of the best-evolved networks, we compute different coverage metrics associated to the dataset corresponding to each possible of the $26$ values of the temperatures considered. We then analyze the changes in these metrics in relation to the temperature. 

   We have used the implementation of the neuron coverage metrics\footnote{Available from \url{https://github.com/DeepImportance/deepimportance_code_release}} developed as part of the work presented in \cite{Gerasimou_et_al:2020}.

\section{Experiments} \label{sec:EXPE}

 In this section, we present the experimental framework and investigate a number of research questions related to the phase detection problems. More specifically, we address the following questions:

  \begin{enumerate}
      \item Whether the NAS procedure is able to generate networks that accurately identify the temperature.
      \item To what extent is possible to identify the phase transition from the predictions made by the best architectures found by NAS?
      \item Whether the coverage metrics are able to capture the system transitions. 
  \end{enumerate}
  
\subsection{Experimental framework}

  Table~\ref{fig:PARAMS} shows the hyperparameters of the evolved neural networks and the parameters that define the neuroevolutionary algorithm. While the number of layers and number of neurons in each layer can vary between neural networks, a maximum of $20$ was set for both values. The maximum number of architectures evaluated in each run of the neuroevolutionary algorithm  was $40 \times 50 =2000$. The algorithm was run $30$ times and the statistical analysis was conducted based on the average metrics computed from these experiments.

 \begin{table}
   \begin{center}      
    \begin{tabular}{|r|r||r|r|} \hline
         max. layers &  20  & Generations & 40  \\  \hline
         max. neurons & 20  & Pop. size &   50   \\ \hline
         n. epochs  &  10    & Train size & 26000  \\  \hline
         batch size & 100   & Val. size  & 13000  \\ \hline
         gradient alg. & Adam &Test size & 26000  \\  \hline                    
    \end{tabular}
    \caption{Hyperparameters of the neural networks and  parameters of the neuroevolutionary algorithm.}
    \label{fig:PARAMS}
    \end{center} 
\end{table}



 \subsection{Performance of the evolved networks}
 
   \begin{figure}[htbp]
    \begin{center}
   \includegraphics[width=8.5cm]{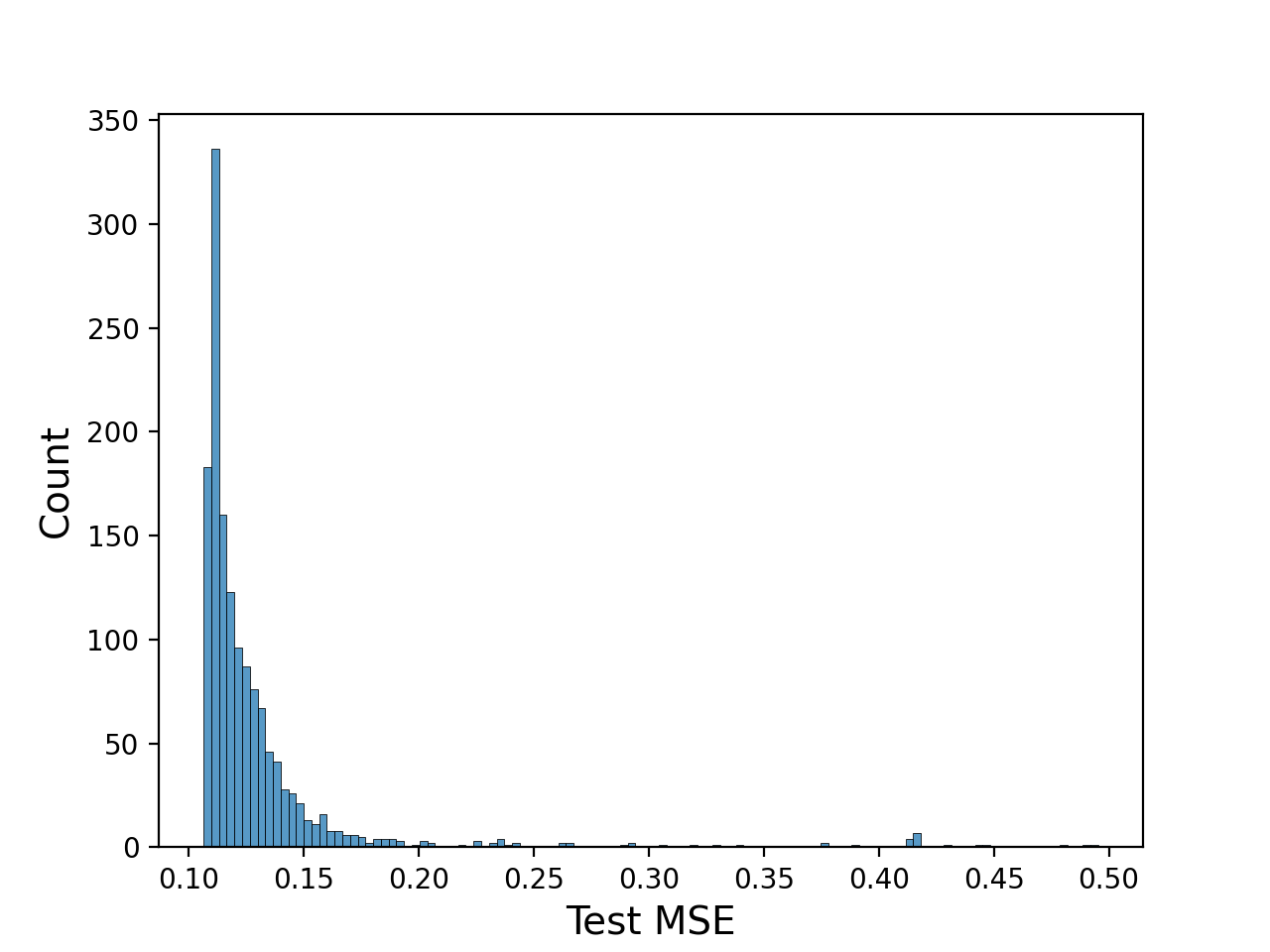}     
   \caption{Histogram of the Test MSE for over $95\%$ of architectures in the last population.}
  \label{fig:HIST_TEST_MSE}
 \end{center}
  \end{figure}

 We evaluate all architectures in the last generation for the $30$ experiments. Of the $1500$ architectures, $1430$ produced predictions with Test MSE below $0.5$. The distribution of the MSE errors for these models are shown in Figure~\ref{fig:HIST_TEST_MSE}. The MSE values corresponding to the other $70$ architectures are not shown for the sake of a clearer visualization. The analysis of the figure reveals that most of the learned architectures are able to produce approximations with a low MSE.


  
 \subsection{Assessing the identification of phase transition from the networks predictions}

  We compute the average MSE of the solutions predicted by the best model of each of the $30$ runs. The errors are computed independently for the datasets corresponding to each of the $26$ values of the temperatures. We would like to test whether the accuracy of the predictions is related to the temperature values.  
  
  \begin{figure}[htbp]
    \begin{center}
   \includegraphics[width=8.5cm]{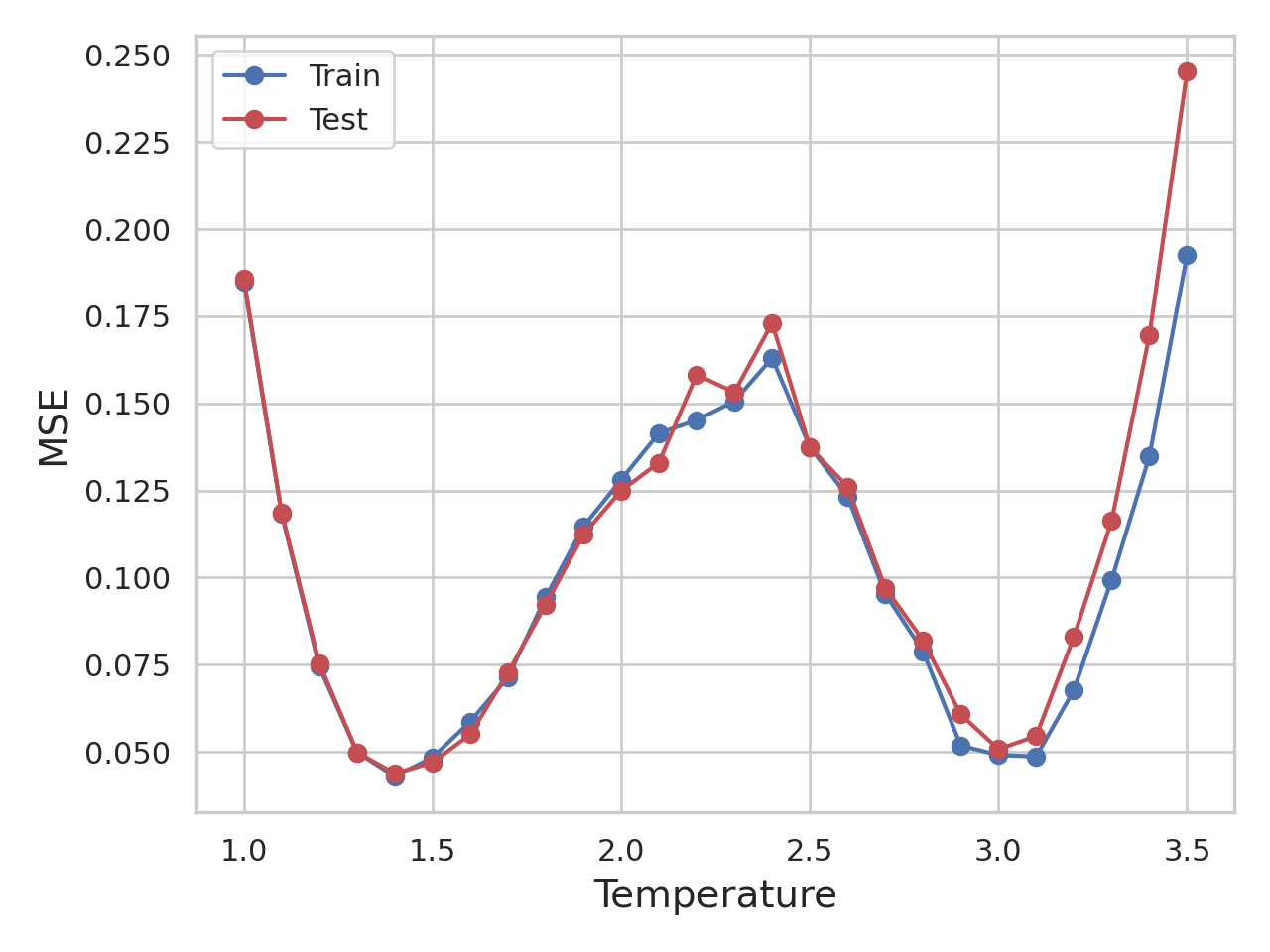}     
   \caption{Mean MSE of the best $30$ models as a function of the temperature.}
  \label{fig:SCORE_BY_TEMP}
 \end{center}
  \end{figure}

    \begin{figure}[htbp]
    \begin{center}
   \includegraphics[width=8.5cm]{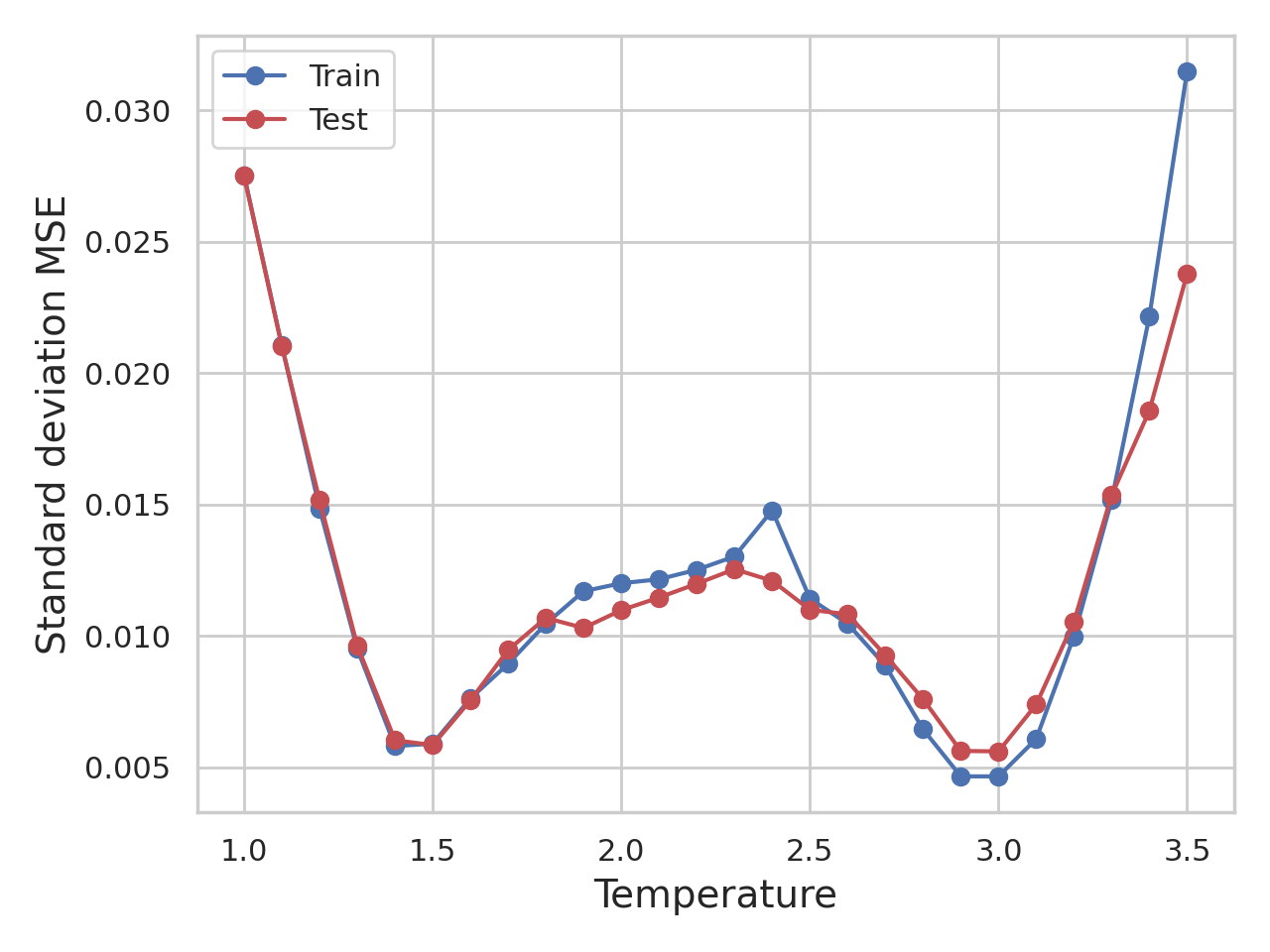}     
   \caption{Standard deviation of MSE of the best $30$ models as a function of the temperature.}
  \label{fig:STD_BY_TEMP}
 \end{center}
  \end{figure}

Figure~\ref{fig:SCORE_BY_TEMP} shows the mean MSE  of the approximations  as a function of the temperature, computed for the train and test sets. It can be seen that the errors are very high for the extreme values of the temperature and also around the critical temperature. The shape of the curves resembles the universal W-shape that has been reported when investigating the learning-by-confusion method \cite{Van-Nieuwenburg_et_al:2017} but in our case making precise predictions for the configurations sampled from the lowest and highest temperatures seems a very hard task for the neural networks. The standard deviation of the MSE, as shown in Figure~\ref{fig:STD_BY_TEMP} reveals a similar pattern. Both, the mean error and the standard deviation are high around the critical temperature.

 \subsection{Can coverage metrics capture phase transitions?}

 For each of the $30$ best architectures, we computed the NC, TKNC, KMN, NBC, and SNAC metrics for each of the $26$ temperatures. Columns 2-3 in Table~\ref{tab:Coverage_Correlation} show the correlation between the coverage metrics and the mean  ($MSE_{mean}$) and standard deviation ($MSE_{std}$) of the MSE of the networks. In column 4, the  correlation between the coverage metrics and the temperature is also displayed. Columns 5-7 present a similar analysis but using the Kendall-tau metric of rank correlation.

\begin{table}
\begin{center}
 \begin{tabular}{lrrr|rrr} \hline
                      &   \multicolumn{3}{c}{Pearson}       &   \multicolumn{3}{c}{Kendall tau}  \\\hline
 Metric               &    MSE &   MSE &   Temp &   MSE &   MSE &   Temp \\
                      &    mean &  std &        &   mean &  std &      \\ \hline
 NC                     &      -0.09 &     -0.65 &   0.52 &       0.33 &      0.04 &   0.17 \\ 
 TKNC                   &       0.29 &     -0.18 &   0.93 &       0.17 &     -0.02 &   0.92 \\
 KMN                    &       0.40 &     -0.11 &   0.72 &       0.42 &      0.22 &   0.48 \\
 NBC                    &       0.44 &      0.27 &   0.77 &       0.21 &      0.02 &   0.62 \\
 SNAC                   &       0.41 &      0.22 &   0.78 &       0.19 &      0.00 &   0.61 \\ 
\hline
\end{tabular}
  \caption{Correlation between the coverage metrics, the mean and standard deviation of MSE, and the temperature.}
\label{tab:Coverage_Correlation}
\end{center}
\end{table}

  \begin{figure}[htbp]
    \begin{center}
   \includegraphics[width=8.5cm]{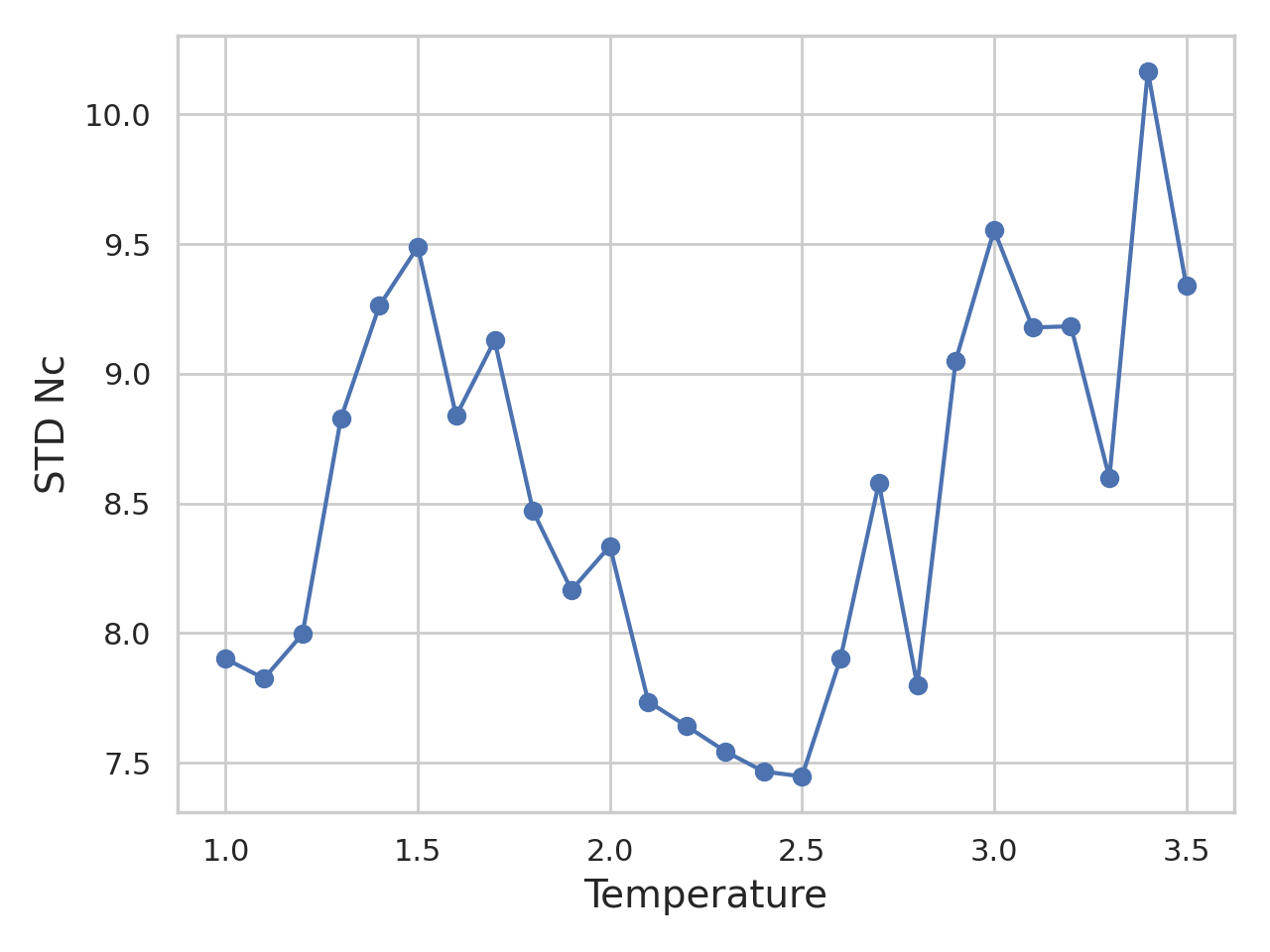}     
   \caption{Standard deviation of the NC metric for the different temperatures.}
  \label{fig:NC_BY_TEMP}
 \end{center}
  \end{figure}

\begin{figure}[htbp]
    \begin{center}
   \includegraphics[width=8.5cm]{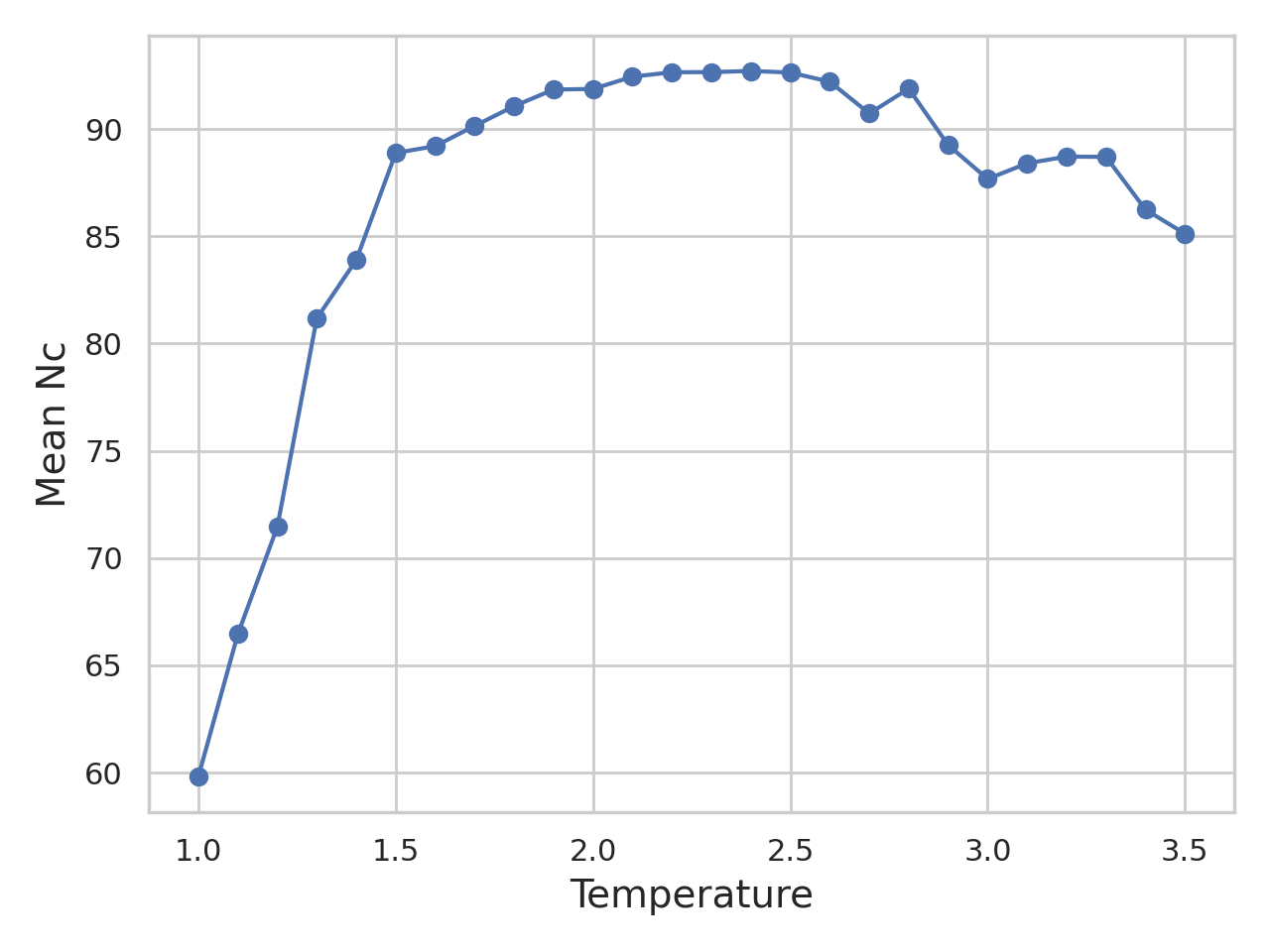}     
   \caption{Mean value of the NC metric for the different temperatures.}
  \label{fig:MEAN_NC_BY_TEMP}
 \end{center}
  \end{figure}
  
 The most remarkable result in the table is the strong negative correlation ($-0.65$) between the standard deviation of the NC metric and the mean MSE. This can be corroborated by comparing Figure~\ref{fig:NC_BY_TEMP} and Figure~\ref{fig:SCORE_BY_TEMP}.  The low variance of the NC metric around the transition region seems to indicate that the number of neurons that are activated for these values of the temperature in these high-performing models is very stable. By analyzing the mean value of the NC metric (Figure~\ref{fig:MEAN_NC_BY_TEMP}),  it can be seen that it is around the critical temperature where the number of activated neurons is maximized. While the values remain high for the remaining temperatures, the combination of a high mean value with a low accuracy point to the possibility that indeed the NC metric captures changes in the network response as a result of the more disordered configurations of the Ising model in these region of the temperatures.

\begin{figure}[htbp]
    \begin{center}
   \includegraphics[width=8.5cm]{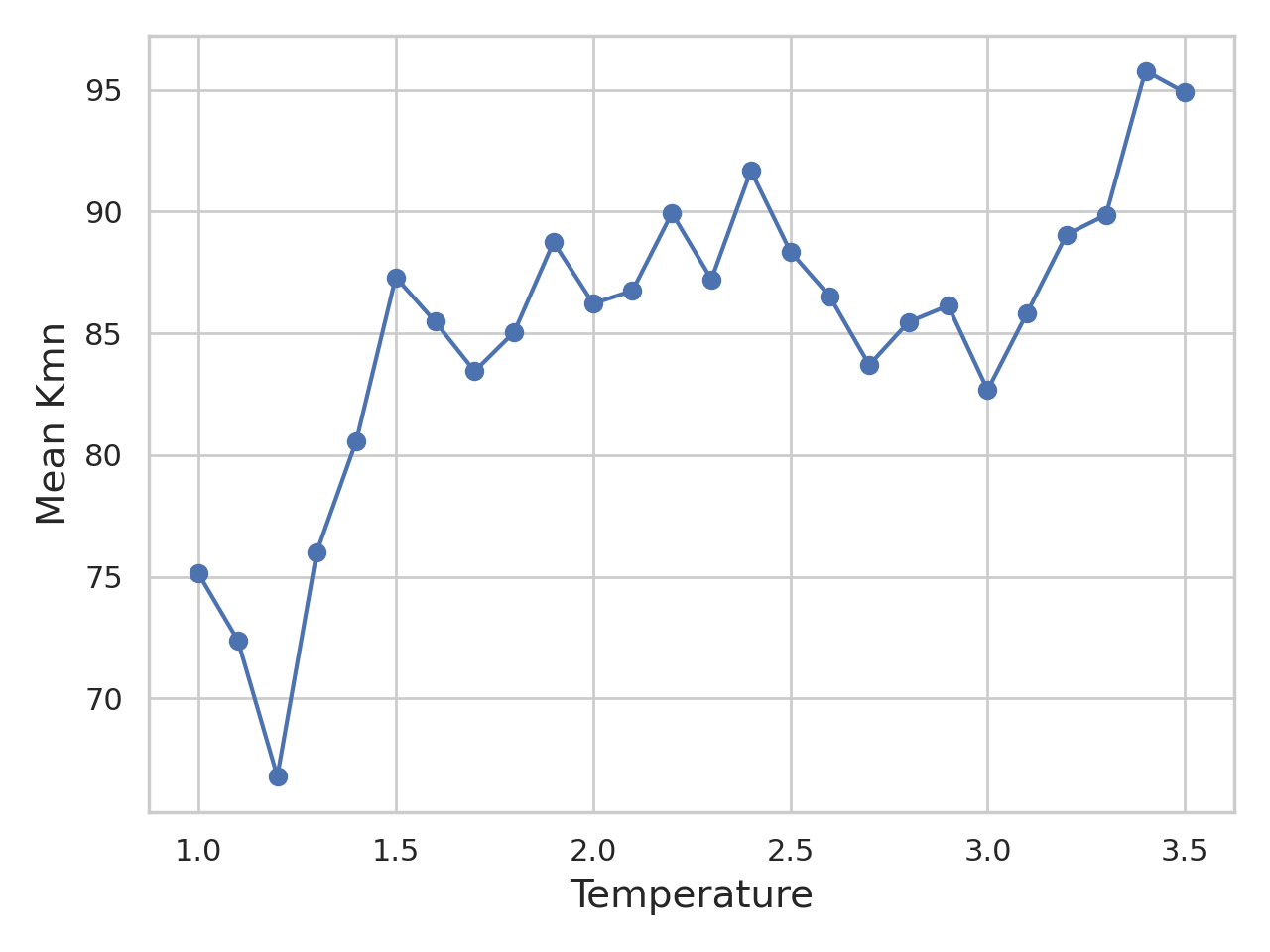}     
   \caption{Mean value of the KMN metric for the different temperatures.}
  \label{fig:KMN_BY_TEMP}
 \end{center}
  \end{figure}

\begin{figure}[htbp]
    \begin{center}
   \includegraphics[width=8.5cm]{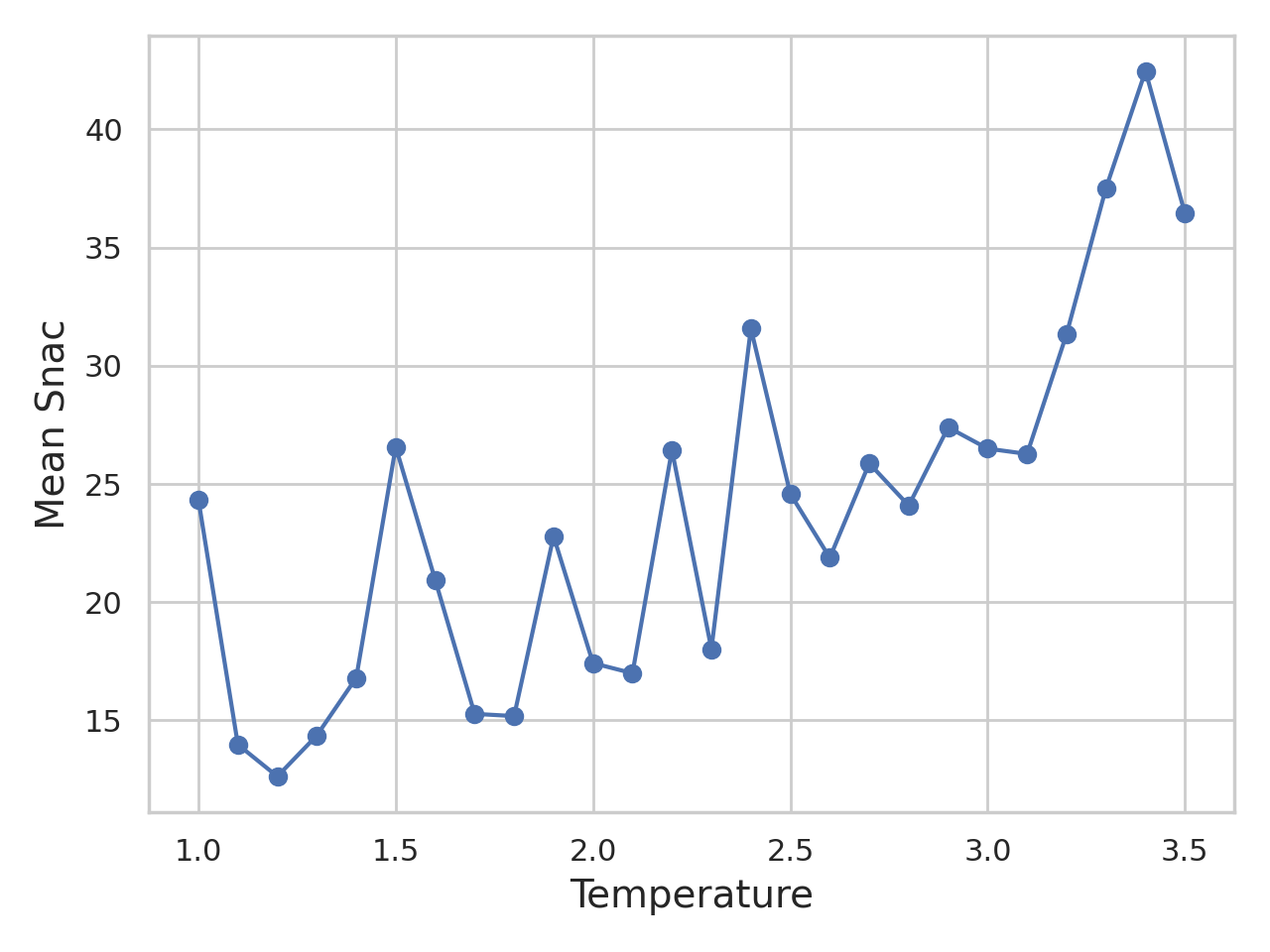}     
   \caption{Mean value of the SNAC metric for the different temperatures.}
  \label{fig:SNAC_BY_TEMP}
 \end{center}
  \end{figure}

 The strong correlations between the coverage  metrics and the temperature, as shown in Table~\ref{tab:Coverage_Correlation} indicates that as the target value that the network has to predict increases, also the coverage metrics increase. This pattern can be appreciated in figures~\ref{fig:KMN_BY_TEMP} and~\ref{fig:SNAC_BY_TEMP}. However, there is also a positive correlation between KMN, NBC and SNAC and the mean MSE. This correlation is above $0.4$ in this case
 
\section{Conclusions} \label{sec:CONCLU}

 The analysis of phase transition is an important problem in Physics and the use of machine learning algorithms has provided new tools for the study of these phenomena. In this paper we have proposed the combined use of neural architecture search and neuron coverage metrics as part of supervised machine learning approaches to investigate phase transition. NAS contributes to create more accurate models that can better capture the mapping between the system configuration and the variable of interest. Coverage metrics provide a different perspective on the performance of the neural networks for the prediction tasks. 
 
  Our results show that the prediction task is harder at the extreme of the range of temperatures analyzed and around the critical temperature. On the other hand, the NC metric stands out as a possible indicator of phase transition with a strong correlation with the mean MSE. 
   
 \subsection{Future work}

 There are a number of ways in which this work could be further developed. Different, more complex, physical systems should be investigated using the approach proposed in this paper. Similarly, both, NAS and the coverage metrics could be used with other supervised ML approaches for phase transition as those based on multi-class classification and learning-by-confusion. This would require to define appropriate optimization objectives for NAS, consistent with the physical systems and the supervised ML approaches. 
  
In the quantum systems viewpoint, our neural network approaches are promising for identifying topological quantum systems and complex many-body properties like entanglement entropy, as suggested in \cite{Torlai_2018}. Extending the investigation to quantum chromodynamics (QCD) at finite temperatures, they could offer novel insights into the confined-deconfined phase transition, aiding in understanding quarks and gluons under extreme conditions by using data from lattice simulations and experiments such as LHC (Large Hadron Collider) and RHIC (Relativistic Heavy-Ion collider) \cite{Nagle:2011uz}, these networks could learn and predict in the unseen data patterns in the QCD phase diagram \cite{Shuryak:1996pb}, including phenomena like color superconductivity and chiral symmetry restoration or even novel physical effects. 

\section*{Acknowledgments}

 R. C. Terin acknowledges the Basque Foundation for Science  for financial support through the IKUR strategy program under projects $214023$FMAO and $214021$ELCN. R. Santana acknowledges funding from the Spanish Ministry of Science, Innovation and Universities (project PID$2022-137442$NB-I$00$, and the Basque Government (projects KK-$2023/00012$ and IT$1504-22$).

\bibliographystyle{IEEEtran}    
\bibliography{phase_transition}

\end{document}